# Bit Error Tolerance Metrics
# for Binarized Neural Networks

Sebastian Buschjäger*, Jian-Jia Chen, Kuan-Hsun Chen, Mario Günzel, Katharina Morik, Rodion Novkin,
Lukas Pfahler*, Mikail Yayla*
*These authors contributed equally
Technical University of Dortmund
{mikail.yayla, lukas.pfahler, rodion.novkin, katharina.morik, mario.guenzel,
kuan-hsun.chen, jian-jia.chen, sebastian.buschjaeger}@tu-dortmund.de

*Abstract*—To reduce the resource demand of neural network (NN) inference systems, it has been proposed to use approximate memory, in which the supply voltage and the timing parameters are tuned trading accuracy with energy consumption and performance. Tuning these parameters aggressively leads to bit errors, which can be tolerated by NNs when bit flips are injected during training. However, bit flip training, which is the state of the art for achieving bit error tolerance, does not scale well; it leads to massive overheads and cannot be applied for high bit error rates (BERs). Alternative methods to achieve bit error tolerance in NNs are needed, but the underlying principles behind the bit error tolerance of NNs have not been reported yet. With this lack of understanding, further progress in the research on NN bit error tolerance will be restrained.

In this study, our objective is to investigate the internal changes in the NNs that bit flip training causes, with a focus on binarized NNs (BNNs). To this end, we quantify the properties of bit error tolerant BNNs with two metrics. First, we propose a neuron-level bit error tolerance metric, which calculates the margin between the pre-activation values and batch normalization thresholds. Secondly, to capture the effects of bit error tolerance on the interplay of neurons, we propose an inter-neuron bit error tolerance metric, which measures the importance of each neuron and computes the variance over all importance values. Our experimental results support that these two metrics are strongly related to bit error tolerance.

## I. Introduction

In recent years, neural networks (NNs) have been applied successfully in various fields. However, NNs rely on a large number of parameters to achieve high accuracy, which leads to high resource-demand, especially for the memory used in the system. It has been reported that the memory uses a large fraction of the entire system energy in NN-based inference systems [3], [7].

To reduce the energy consumption of the memory, recent studies have proposed to exploit approximate memory, by tuning the memory supply voltage and timing parameters of NN-based inference systems. If these approaches are applied aggressively, bit errors can occur. For modern memory technologies, such as volatile memories (SRAM [13], DRAM [7]) and emerging non-volatile memories (e.g. STT-RAM [5], [11], RRAM [4]) the bit error rate (BER) increases steeply in these approaches. These bit errors can potentially degrade the accuracy of NNs to unacceptable levels if no countermeasures are employed.

Huitzil et al. [12] provide a thorough overview of the recent research results in error tolerant NNs. To achieve error tolerance in NNs, Edwards et al. [2] explore a penalty term for optimal distribution of computations to neurons. Cavalieri et al. [1] propose to distribute to the neurons the absolute values of weights evenly. Currently, the state of the art for achieving bit error tolerance in NNs is bit flip injections during training [4], [7].

However, bit flip training has two main disadvantages which limit its usage in practice. First, injecting bit flips during training can lead to significant accuracy degradation, which becomes more severe the higher the BER during training [4], [7]. Secondly, it leads to serious additional overhead, since the bit error model has to be applied during the training process [9]. Therefore, it is necessary to research alternative methods to achieve bit error tolerance, without bit flip injection during training.

The theoretical foundations of NN bit error tolerance, however, have not been reported yet. With this lack of knowledge, further research on the bit error tolerance of NNs may be hindered. Understanding the underlying principles of bit error tolerance is necessary to gain new insights and to create novel, more powerful methods for the bit error tolerance of NNs, to enable lower power consumption for NN-based inference.

In this study, our goal is to gain understanding of the underlying principles behind the bit error tolerance in NNs. To this end, we investigate two metrics regarding the internal changes in the NNs that bit flip training causes: (1) The neuron-level bit error tolerance metric, which captures the effect of bit error tolerance on individual neurons, and (2) the inter-neuron bit error tolerance metric, which captures the effect on the interplay between the neurons. In our investigations, we focus on binarized neural networks (BNNs), which are highly robust against bit errors [4] and have a simpler structure than NNs with higher precision weights.

This paper has been supported by Deutsche Forschungsgemeinshaft (DFG) as part of the Collaborative Research Center SFB 876 "Providing Information by Resource-Constrained Analysis" (project number 124020371) and project OneMemory (project number 405422836).

**Our contributions:**

- We define a neuron-level bit error tolerance metric for BNNs in Section III and formally prove that it measures the number of bit flips a neuron can tolerate. This bit error tolerance metric is constructed without bit flip injections, by calculating the margin between the batch normalization thresholds and the pre-activation values.
- To capture the effects of bit flip training on the interplay of neurons, in Section IV, we investigate an inter-neuron bit error tolerance metric, which is defined as the variance of a neuron importance metric. We propose to estimate the importance of individual neurons by flipping their outputs and subsequently determining the accuracy degradation.
- In our experiments in Section V, we first show the correlation between accuracy over BER and the neuron-level bit error tolerance metric. This indicates that on the neuron-level, *bit flip training in BNNs leads to large margins between the batch normalization thresholds and pre-activation values*. Moreover, for smaller BNNs, we experimentally show that the inter-neuron bit error tolerance metric is related to bit error tolerance. This indicates that on the inter-neuron-level, *bit flip training in BNNs leads to a low variance of neuron importance values* for small BNNs.

## II. SYSTEM MODEL

Each layer in a convolutional or fully connected neural network (NN) with floating point or integer weights receives an input and computes a convolution between it and the weights. The result of the activation is fed to the subsequent layer and this process is repeated until the final output is computed. This procedure is called the *forward pass*.

In binarized NNs (BNNs), a resource-efficient variant of NNs, the weights and activations are binarized. In this case, the output of a layer can be computed with

$$2 * popcount(xnor(W_i^l, X^{l-1})) - \#bits > T,$$

where $popcount$ counts the number of 1s in the xnor-result, $\#bits$ is the number of bits in the $xnor$ operands, and $T$ is a learnable threshold parameter (computed with the batch normalization parameters), whose comparison produces binary values (representing a shifted binarization function) [6], [8].

### A. Training

To train NNs, a common method is applying stochastic gradient descent (SGD) with mini-batches. In this work, the training data is described with $\mathcal{D} = \{(x_1, y_1), \ldots, (x_I, y_I)\}$ with $x_i \in \mathcal{X}$ as the inputs, $y_i \in \mathcal{Y}$ as the labels, and $\ell \colon \mathcal{Y} \times \mathcal{Y} \to \mathbb{R}$ as the loss function. $W = (W^1, \ldots, W^L)$ are the weight tensors of layer $1 \ldots L$ and $f_W(x)$ is the output of the NN. The goal is to find a solution for the optimization problem

$$\arg\min_W \frac{1}{I} \sum_{(x,y) \in \mathcal{D}} \ell(f_W(x), y)$$

by a mini-batch SGD strategy by computing gradients using backpropagation. To train BNNs, Hubara et al. [6] proposes to deterministically binarize the weights and activations during the forward pass. But for backpropagation, the floating point numbers are used for parameter updates. To make BNNs bit error tolerant, the state-of-the-art method is bit flip injections in the binarized values during forward pass, as proposed by Hirtzlin et al. [4].

### B. BNN Building Blocks

In this work, we consider two types of object recognition NNs: fully connected binarized neural networks (FCBNNs) and convolutional binarized neural networks (CBNNs). The NNs we use are build up of the following layer types: 1) convolutional, 2) maxpool, 3) batch normalization followed by activation, and 4) fully connected layer.

1) The convolutional layer computes a 2D convolution of the input with a specified number of filters, e.g. with 64 we write C64. C layers have binary weights as parameters, which can be stored in one bit. In this work, we use filters of size $3 \times 3$ only. 2) The maxpool layer downsamples the input by selecting the maximum value of the input in a given window size, with 2 we write MP2. MP has no trainable parameters. We use maxpool with window size $2 \times 2$ only in this work. 3) The batch normalization (BN) layer is used for faster and more stable training. In this work, it is always followed by the binary activation function. For inference, the BN layer followed by activation can be computed by binary thresholding [8]. BN layers have thresholds as parameters, which are signed integers. 4) The fully connected layer connects every neuron in the current layer with every neuron in the next layer. We write FC1024 when 1024 neurons are used to compute the output of the FC layer. FC layers also have binary weights as learnable parameters.

### C. Memory model

In this study, we assume transient and symmetric bit error rates (BERs), which means the probability to read a 0 instead of a 1 is the same as the probability to read a 1 instead of a 0. A bit flip is reversed after the content of the memory cell has been read from approximate memory. The effects of the bit flips can, however, propagate to the subsequent layers of the NNs. This memory model is in line with models that are assumed in recent studies, which explore approximate volatile memories (SRAM [13], DRAM [7]), and approximate non-volatile memories (RRAM [4], MRAM or STT-RAM [5]).

## III. NEURON-LEVEL BIT ERROR TOLERANCE METRIC

To understand the bit error tolerance of BNNs, we propose to formalize it using a metric that is calculated on the neuron level. To this end, our presentation in this section focuses on convolutional neural networks (CNNs). We note that all of our definitions are also valid and applicable to fully connected neural networks (FCNNs), by changing the filter size. (In our experiments, both CNNs and FCNNs were evaluated in Section V.) We first define the bit error tolerance of a $2d$ feature map in a CNN. Then we leverage this definition into the bit error tolerance of a single neuron, which enables us to define a metric for the bit error tolerance of the entire NN.

Consider a CNN and let $n$ be the index of one neuron. The output of the neuron is a 2d feature map with height $U$ and width $V$. Here, we consider the neuron as a function, which slides a filter over an input tensor. We define the neuron's *bit error tolerance* $T_{i,n,u,v}$ at position $u,v$ by the number of weight sign flips it can tolerate without a change of its output given the input $x_i$. Let $h_{i,n,u,v} \in \mathbb{Z}$ be the output of neuron $n$ *before* applying the activation function. For neurons that are not in the first layer, we note two points: First, each neuron's output is computed by a weighted sum of inputs that are $\pm 1$ with weights that are also $\pm 1$. Second, the sign function is applied to this output. Thus, as long as weight flips do not change the sign of the weighted sum, a neuron is bit error tolerant. We can quantify the bit error tolerance of a neuron given the input $x_i$ by the distance of its output from 0:

$$T_{i,n,u,v} = \left| h_{i,n,u,v} - s_n - \tfrac{1}{2} \right| \tag{1}$$

We include $s_n \in \mathbb{Z}$ to account for activation shifts due to the batch normalization layer (without $s_n = 0$), and to avoid ambiguity at 0 we subtract $\tfrac{1}{2}$. In this equation, the higher the difference between $h_{i,n,u,v}$ and $s_n - \tfrac{1}{2}$, the higher $T_{i,n,u,v}$. Note that $T_{i,n,u,v}$ is a measure for the worst-case error tolerance, in the sense that at least $\left\lfloor \frac{T_{i,n,u,v}}{2} \right\rfloor + 1$ weight sign flips are necessary to change the output sign. With each weight sign flip $h_{i,n,u,v}$ can get closer to $s_n$ and finally flip the output. For neurons in the first layer, the inputs are not in $\{\pm 1\}$ but $\{0, \ldots, Z\}$, thus we scale the bit error tolerance:

$$T_{i,n,u,v} = \frac{\left| h_{i,n,u,v} - s_n - \tfrac{1}{2} \right|}{Z} \tag{2}$$

The definition of $T_{i,n,u,v}$ yields the following theorem:

**Theorem 1.** *Let $b \in \mathbb{R}_{\geq 0}$. If $T_{i,n,u,v} \geq b$ for all $u,v$ then the neuron can tolerate at least $\lfloor \tfrac{b}{2} \rfloor$ bit flips, i.e. any bit flip of up to $\lfloor \tfrac{b}{2} \rfloor$ weights of the neuron does not affect its output.*

*Proof.* At first, we consider the neuron $n$ and assume that it is not in the first layer. Let $u,v$ be a position for the convolution result. As described in (1), we know that $T_{i,n,u,v} = |h_{i,n,u,v} - s_n - \tfrac{1}{2}|$. For improved readability, we write $h$ for $h_{i,n,u,v}$ and $s$ for $s_n$. By construction of the activation function, the output of the neuron at position $u,v$ is $+1$ for $h - s - \tfrac{1}{2} > 0$ and $-1$ for $h - s - \tfrac{1}{2} < 0$ since $h$ and $s$ are assumed to be integer values. Furthermore, the case $h - s - \tfrac{1}{2} = 0$ does not occur. If $T_{i,n,u,v} \geq b$ then either $h - s - \tfrac{1}{2} \geq b$ or $h - s - \tfrac{1}{2} \leq -b$.

In the first case we have $h - s - \tfrac{1}{2} \geq b \geq 0$ and the output at $u,v$ is $+1$. We denote by $\tilde{h}$ the value of $h$ after the bit flips of up to $\lfloor \tfrac{b}{2} \rfloor$ weights. By definition, $h$ is a weighted sum where each summand is one input of the neuron multiplied with one weight. Since each summand is in $\{\pm 1\}$, changing one sign changes $h$ by 2. Therefore $\tilde{h}$ can differ by up to $2 \cdot \lfloor \tfrac{b}{2} \rfloor$ from $h$ and is in $[h - \lfloor b \rfloor, h + \lfloor b \rfloor]$. For $\tilde{h}$ we still have $\tilde{h} - s - \tfrac{1}{2} \geq b - \lfloor b \rfloor \geq 0$ which causes an output of $+1$ at position $u,v$ of the neuron.

The second case is proven analogously: This time we assume that $h - s - \tfrac{1}{2} \leq -b \leq 0$. Then the output of the neuron at $u,v$ is $-1$. Changing the sign of $\lfloor \tfrac{b}{2} \rfloor$ summands of $h$ can increase the value of $h$ by up to $\lfloor b \rfloor$. After at most $\lfloor \tfrac{b}{2} \rfloor$ bit flips, we obtain a new value $\tilde{h} \in [h - \lfloor b \rfloor, h + \lfloor b \rfloor]$. We have $\tilde{h} - s - \tfrac{1}{2} \leq -b + \lfloor b \rfloor \leq 0$ and the output of this neuron at $u,v$ is still $-1$.

If neuron $n$ is in the first layer, then we have $T_{i,n,u,v} = \frac{|h-s-\tfrac{1}{2}|}{Z}$ by (2). Since the summands of $h$ are in $[-Z, Z]$, the value of $h$ can change by up to $2 \cdot Z$ per bit flip. Thus, after $\lfloor \tfrac{b}{2} \rfloor$ bit flips the value of $\tilde{h}$ is in the interval $[h - Z \cdot \lfloor b \rfloor, h + Z \cdot \lfloor b \rfloor]$. But the proof still works as above since $|h - s - \tfrac{1}{2}| \geq Z \cdot b$ by assumption. $\square$

We note that the theorem can be extended to include bit flips in the binary input values of the neuron as well. This would not change the theorem, because of the XNOR operation applied on the input and weight bits. The flip of either the input or the weight leads to a flip of the XNOR result, therefore the number of bit flips (input and weight bits) a neuron can tolerate is the same as the number of weight flips it can tolerate. In this consideration, we exclude the input values of the first layer.

Intuitively, a neuron is bit error tolerant if it is robust across all positions. We demand that each position has a bit error tolerance of at least $b$. Formally, the *bit error tolerance* $T_{i,n}^b$ of a neuron $n$ given the input $x_i$ is defined as:

$$T_{i,n}^b = \frac{1}{UV} \sum_{u=1}^{U} \sum_{v=1}^{V} \mathbb{1}\{T_{i,n,u,v} \geq b\} \tag{3}$$

The bit error tolerance of the entire network can then be defined as the average error tolerance across all neurons:

$$T_i^b = \frac{1}{N} \sum_{n=1}^{N} T_{i,n}^b \tag{4}$$

We determine $T^b$ by evaluating the BNN on the entire dataset:

$$T^b = \frac{1}{I} \sum_{i=1}^{I} T_i^b \tag{5}$$

For a set of values $b \in \{b_1, \ldots, b_B\}$ we consider $T$ to be the tuple $T = (T^{b_1}, \ldots, T^{b_B})$. For different neural networks we use $T$ to compare their neuron-level bit error tolerance.

## IV. INTER-NEURON BIT ERROR TOLERANCE METRIC

The metric in Section III is constructed based on the properties of individual neurons with the same importance, which by definition cannot perfectly capture the interplay between neurons. In this section, our goal is to capture the effect of bit flip training on the interplay between neurons.

To get an intuition for a metric, we need to reflect upon the way neurons adapt to weight flips. Bit flips in the weights can cause activation flips of neurons, which BNNs need to adapt to while still keeping accuracy high. We hypothesize that in weight flip training, neurons learn adaptations among each other that correct the neuron activation flips. These adaptations

**Algorithm 1:** Neuron Importance Metric $\Pi$

**Input:** $model, (X, y)$
**Output:** $\Pi$

1 $\Pi = \{\Pi[0], \Pi[1], \dots\}$
2 $A = accuracy(model(X), y)$
3 **for** *each neuron $n$* **do**
4 $\quad$ flip the activation output of neuron $n$
5 $\quad$ $A^* = accuracy(model(X), y)$
6 $\quad$ $\Pi[n] = \frac{A - A^*}{A}$

may be internal correction mechanisms, such as redundancies or distribution of computation. Here, we propose to quantify these adaptations by estimating the dispersion of the individual accuracy-importance of neurons.

To achieve this, we propose the neuron importance metric, denoted as $\Pi[n]$, which measures the importance of neuron $n$. With $A$ as the accuracy without bit flips and $A^*$ the accuracy when the output of neuron $n$ is flipped, the neuron importance of individual neurons is defined as the accuracy drop $\Pi[n] = \frac{A - A^*}{A}$, i.e. normalized to the accuracy without bit flips.

We calculate $\Pi[n]$ for all neurons with Algorithm 1. The algorithm takes a trained model as input, and the metric is computed for the entire test set. $\Pi$ in Line 1 is an array which holds neuron importance values for every neuron in the BNN. In Line 2, the accuracy $A$ on the test set without bit flips is evaluated. From Line 4 on, for every neuron $n$ the accuracy-importance is determined on the entire test set. The algorithm first determines the accuracy $A^*$ on the test set when flipping only the output of the neuron under consideration. Then $A^*$ is subtracted from $A$, and the result is divided by $A$ to finally acquire the neuron importance value $\Pi[n]$. We measure the neuron importance of every neuron in the BNN, except for neurons in the output layer.

To measure the scattering of $\Pi[n]$, we calculate the variance of all neuron importance values

$$\text{VAR}(\Pi) = \frac{1}{N}\sum_{n=1}^{N}(\Pi[n] - \overline{\Pi})^2, \quad (6)$$

which is our inter-neuron bit error tolerance metric. The mean of $\Pi[n], n = 1, \dots, N$ is $\overline{\Pi} = \frac{1}{N}\sum_{n=1}^{N}\Pi[n]$, and $\Pi[n] - \overline{\Pi}$ is the difference between one neuron importance value and the mean of all neuron importance values of the BNN. We hypothesize that redundancy and distribution of computation results in low variance of neuron importance as described in the following hypothesis:

**Hypothesis 1.** *BNNs trained with weight flips (as described in Section II) have a lower variance of the neuron importance values when compared to BNNs trained without weight flips.*

V. EXPERIMENTS

| Name | # Train | # Test | # Dim | # classes |
|---|---|---|---|---|
| FashionMNIST | 60000 | 10000 | (1,28,28) | 10 |
| CIFAR10 | 50000 | 10000 | (3,32,32) | 10 |

TABLE I: Datasets used for experiments.

| Parameter | Range |
|---|---|
| Flip probability | $p \in \{1\%, 5\%, 10\%, 20\%\}$ |
| Fashion FCNN | In $\to$ FC 2048 $\to$ FC 2048 $\to$ 10 |
| Fashion CNN | In $\to$ C64 $\to$ MP 2 $\to$ C64 $\to$ MP 2 $\to$ FC2048 $\to$ 10 |
| CIFAR10 CNN | In $\to$ C128 $\to$ C128 $\to$ MP 2 $\to$ C256 $\to$ C256 $\to$ MP 2 $\to$ C256 $\to$ C256 $\to$ MP 2 $\to$ FC 2048 $\to$ 10 |

TABLE II: Parameters used for experiments.

We first present the experiment setup in Section V-A. In Section V-B, we provide experimental support for the neuron-level error tolerance metric and in Section V-C we do the same for Hypothesis 1 posed in Section IV.

*A. Experiment Setup*

We evaluate fully connected binarized neural networks (FCBNNs) and convolutional binarized neural networks (CBNNs) in the configurations shown in Table II for the datasets FashionMNIST and CIFAR10 (see Table I). In all experiments, we run the Adam optimizer for 100 epochs for FashionMNIST and 250 epochs for CIFAR10 to minimize the cross entropy loss. We use a batch size of 128 and an initial learning rate of $10^{-3}$. To stabilize training, we exponentially decrease the learning rate every 25 epochs by 50 percent.

*B. Neuron-level Bit Error Tolerance Metric*

In this subsection, our goal is to find out whether there is experimental support for the neuron-level bit error tolerance metric (proposed in Section IV). We plot the accuracy over BER for BNNs trained *without bit flip injection* (No flips) and *with bit flip injection* (Flip, $p$), in the top row of Fig. 1. In the bottom row, we plot the results for the neuron-level bit error tolerance metric $T^b$, with $b = \{2, 4, 8, 16, 32, 64\}$.

For all three cases tested, we observe that the accuracy over different BERs correlates with $T^b$. When a BNN achieves high bit error tolerance, we observe high $T^b$. We note that the difference between the $T^b$ curves is higher for Fashion than for CIFAR10. Furthermore, the higher the BERs during training, the higher the values of the $T^b$ curve.

*C. Inter-neuron and Neuron-level Bit Error Tolerance Metrics*

In this section, we evaluate the inter-neuron bit error tolerance metric $\text{VAR}(\Pi)$ to gain insight into the effects of bit error tolerance on the interplay of neurons. The goal is to evaluate Hypothesis 1 in Section IV. Furthermore, we want to evaluate the cases in which the inter-neuron or the neuron-level metric explain the bit error tolerance. In the experiments we use instances of the Fashion BNNs in Table I with different number of neurons and train each BNN until convergence.

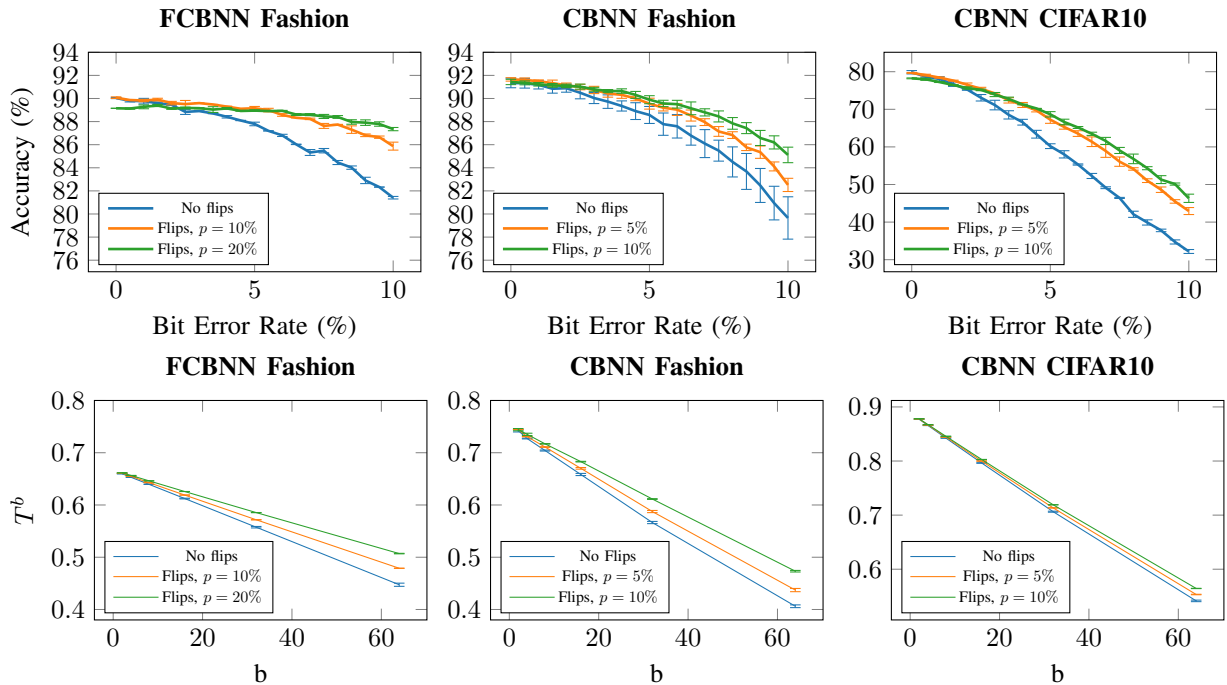

Fig. 1: The relationship between accuracy over BER and $T^b$ with $b = \{2, 4, 8, 16, 32, 64\}$. Top row: accuracy over BER, bottom row: $T^b$ values plotted over $b$. FCBNN means fully connected NN, CBNN means convolutional BNN.

In Fig. 2, we show representative examples of BNNs trained with 5% BER from the cases evaluated in Table IIIa and IIIb. We observe that in BNNs trained with bit flips, the neuron importance values are less dispersed. This supports Hypothesis 1, for the case of smaller FCBNNs and CBNNs. For the case of 64 and more neurons per layer, however, the dispersion does not become smaller. Values below zero (i.e., improving accuracy) are also possible for individual importance-values, this can be observed in Fig. 2 as well.

To evaluate Hypothesis 1 in more detail, we test more FCBNNs in Table IIIa and CBNNs in Table IIIa. We test the following BERs during training: $\{0\%, 1\%, 5\%, 10\%, 20\%, 30\%\}$. For the $\text{VAR}(\Pi)$ experiments, we apply Algorithm 1 in Section IV. We also compare $\text{VAR}(\Pi)$ to the neuron-level bit error tolerance with the summary metric $\widetilde{T}$ for neuron-level bit error tolerance, which is $T$ summed over all its tuple entries and then averaged, here with $b = \{2, 4, 8, 16, 32, 64\}$.

In the experiments, smaller BNNs trained with bit flip injection have a lower average $\text{VAR}(\Pi)$ than BNNs trained without. The variance of the neuron importance is higher for BNNs trained without bit flip injection. This supports Hypothesis 1. However, for larger BNNs (64 neurons and more), the variance is not smaller in bit error tolerant BNNs. In these cases, $T$ shows higher values. For the case with 16 neurons in Table IIIa and 1% BER, neither $\text{VAR}(\Pi)$ nor $T$ explain the bit error tolerance; the BNN may not need to adapt to the bit errors since the BER is low.

In conclusion, for bit error tolerant BNNs, we observe low $\text{VAR}(\Pi)$ in smaller bit error tolerant BNNs, while in larger bit error tolerant BNNs we only observe larger $T$. Our results indicate that the larger the BNN, the better we can explain the bit error tolerance with $T$, and the smaller the BNNs, the better $\text{VAR}(\Pi)$ can explain the bit error tolerance.

## VI. CONCLUSION

To understand the internal changes on the NN that bit flip training causes, we focused on BNNs and proposed two metrics. First, with the neuron-level bit error tolerance metric, we captured the effect of bit error tolerance on individual neurons. Second, we explained the effects of bit error tolerance training on the interplay between the neurons with the inter-neuron bit error tolerance metric. Our results indicate that the bit error tolerance can be explained with $T$ for BNNs with a higher number of neurons and for a smaller number of neurons with $\text{VAR}(\Pi)$. We expect that these contributions will pave the way for researchers to create new and more powerful methods for achieving NN bit error tolerance, enabling the possibility for further energy saving. In the future, we plan to evaluate methods which optimize with respect to $T$ or $\text{VAR}(\Pi)$, to achieve bit error tolerance without bit flip injection. These evaluations may also consider to construct the metrics in a different way, e.g. layer-wise and in combination with other known metrics, such as the gradient-based importance-signal described in [10].

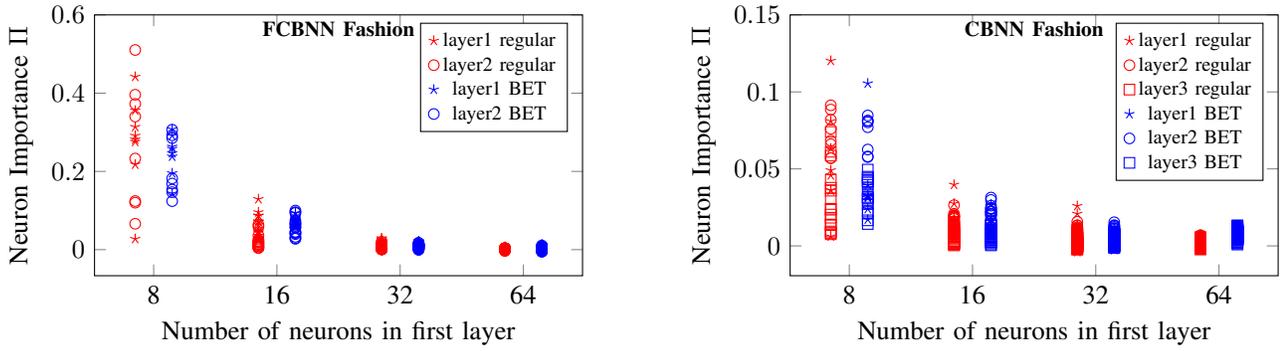

Fig. 2: Representative examples from Table III for neuron importance values Π of FCBNNs (left) and CBNNs (right). Regular BNNs in red, and BNNs trained with bit errors in blue. "BET": trained with 5% BER, "regular": no bit flips during training. X-axis: number of neurons in the first layer. Neurons of different layers have different symbols.

| Neurons | Metric | BER 0% | BER 1% | BER 5% | BER 10% | BER 20% | BER 30% |
|---|---|---|---|---|---|---|---|
| 8,8 | $\widetilde{\Pi}$ | **0.0128** (0.0222,0.00795) | 0.00983 (0.0148,0.00473) | 0.00402 (0.00631,0.00153) | 0.00239 (0.00409,0.00146) | 0.00149 (0.00265,0.000627) | 0.0203 (0.00370,0.000796) |
|  | $\widetilde{T}$ | **0.228** (0.245,0.210) | 0.216 (0.261,0.183) | 0.226 (0.240,0.205) | 0.259 (0.284,0.217) | 0.325 (0.369,0.260) | 0.395 (0.493,0.342) |
| 16,16 | $\widetilde{\Pi}$ | **0.000670** (0.00107,0.000332) | 0.000700 (0.00162,0.000351) | 0.000488 (0.000714,0.000323) | 0.000601 (0.000918,0.000393) | 0.000866 (0.00113,0.000502) | 0.00103 (0.00152,0.000694) |
|  | $\widetilde{T}$ | **0.274** (0.295,0.238) | 0.270 (0.301,0.241) | 0.300 (0.325,0.262) | 0.314 (0.336,0.292) | 0.372 (0.398,0.328) | 0.446 (0.525,0.401) |
| 32,32 | $\widetilde{\Pi}$ | **5.32e-5** (0.000174,1.97e-5) | 3.66e-5 (5.31e-05,2.43e-05) | 3.76e-5 (4.980e-05,2.36e-05) | 5.40e-5 (9.0198e-05,3.35e-05) | 2.50e-4 (0.000305,0.000176) | 6.50e-4 (0.000770,0.000464) |
|  | $\widetilde{T}$ | **0.338** (0.372,0.308) | 0.342 (0.353,0.311) | 0.347 (0.373,0.325) | 0.376 (0.388,0.364) | 0.434 (0.448,0.413) | 0.511 (0.531,0.487) |
| 64,64 | $\widetilde{\Pi}$ | **8.60e-06** (1.37e-05,4.46e-06) | 1.16e-05 (1.82e-05,8.47e-06) | 1.25e-05 (1.57e-05,9.75e-06) | 1.52e-05 (1.99e-05,1.24e-05) | 4.59e-05 (5.40e-05,3.64e-05) | 0.000339 (0.000419,0.000252) |
|  | $\widetilde{T}$ | **0.413** (0.423,0.405) | 0.419 (0.429,0.407) | 0.433 (0.446,0.418) | 0.444 (0.455,0.428) | 0.490 (0.506,0.480) | 0.596 (0.605,0.587) |
| 128,128 | $\widetilde{\Pi}$ | **3.27e-06** (5.07e-06,2.23e-06) | 5.35e-06 (6.32e-06,4.50e-06) | 6.98e-06 (8.00e-06,5.82e-06) | 8.25e-06 (9.90e-06,7.04e-06) | 1.59e-05 (1.79e-05,1.36e-06) | 9.22e-05 (0.000111,8.16e-05) |
|  | $\widetilde{T}$ | **0.502** (0.509,0.491) | 0.508 (0.523,0.498) | 0.519 (0.528,0.510) | 0.532 (0.539,0.525) | 0.558 (0.566,0.550) | 0.647 (0.658,0.638) |

(a) Fully connected BNNs Fashion

| Neurons | Metric | BER 0% | BER 1% | BER 5% | BER 10% | BER 20% | BER 30% |
|---|---|---|---|---|---|---|---|
| 8,8,16 | $\widetilde{\Pi}$ | **0.000991** (0.00272,0.000474) | 0.000924 (0.00162,0.000510) | 0.000662 (0.00114,0.000406) | 0.000827 (0.00126,0.000547) | 0.000867 (0.00154,0.000531) | 0.00107 (0.00175,0.000727) |
|  | $\widetilde{T}$ | **0.341** (0.352,0.333) | 0.344 (0.352,0.329) | 0.359 (0.370,0.351) | 0.379 (0.390,0.369) | 0.439 (0.451,0.423) | 0.540 (0.548,0.519) |
| 16,16,32 | $\widetilde{\Pi}$ | **7.15e-05** (0.000119,4.50e-05) | 6.03e-05 (7.90e-05,4.89e-05) | 6.16e-05 (8.71e-05,5.03e-05) | 6.49e-05 (8.53e-05,5.05e-05) | 0.000151 (0.000232,0.000116) | 0.00048 (0.000650,0.000404) |
|  | $\widetilde{T}$ | **0.414** (0.418,0.411) | 0.418 (0.424,0.411) | 0.434 (0.440,0.429) | 0.450 (0.457,0.446) | 0.492 (0.502,0.481) | 0.574 (0.591,0.554) |
| 32,32,64 | $\widetilde{\Pi}$ | **1.32e-05** (2.26e-05,7.60e-06) | 1.26e-05 (1.76e-05,1.04e-05) | 1.19e-05 (1.42e-05,1.06e-05) | 1.40e-05 (1.77e-05,1.03e-05) | 3.0e-05 (3.62e-05,2.27e-05) | 0.000184 (0.000259,0.000135) |
|  | $\widetilde{T}$ | **0.485** (0.489,0.479) | 0.489 (0.494,0.484) | 0.507 (0.510,0.500) | 0.523 (0.526,0.520) | 0.550 (0.555,0.541) | 0.603 (0.611,0.593) |
| 64,64,128 | $\widetilde{\Pi}$ | **4.02e-06** (5.71e-06,2.85e-06) | 5.19e-06 (5.98e-06,4.45e-06) | 5.71e-06 (6.92e-06,4.96e-06) | 6.24e-06 (6.70e-06,5.78e-06) | 1.06e-05 (1.22e-05,9.47e-06) | 4.29e-05 (5.63e-05,3.64e-05) |
|  | $\widetilde{T}$ | **0.546** (0.549,0.545) | 0.552 (0.555,0.549) | 0.569 (0.571,0.566) | 0.589 (0.590,0.586) | 0.608 (0.611,0.600) | 0.644 (0.647,0.640) |

(b) Convolutional BNNs Fashion

TABLE III: $\widetilde{\Pi}$: average of VAR(Π) over ten experiment repetitions, in brackets: (VAR(Π)$_{max}$, VAR(Π)$_{min}$). $\widetilde{T}$: $T$ summed over its tuple entries and then averaged, in brackets: ($T_{min}, T_{max}$). BERs during training are on top. The nr. of neurons are for the first layers. Blue indicates that Π explains the bit error tolerance, while brown indicates that $T$ explains it.